\documentclass[letterpaper, 10 pt, conference]{ieeeconf}
\IEEEoverridecommandlockouts
\usepackage[compress]{cite}
\makeatletter
\let\NAT@parse\undefined
\makeatother
\usepackage[colorlinks=true]{hyperref}
\usepackage{dblfloatfix}
\usepackage[skip=2pt,font=footnotesize]{caption} %
\usepackage{upgreek,bm}
\usepackage{amsmath, amssymb, amsfonts}
\DeclareMathSymbol{\shortminus}{\mathbin}{AMSa}{"39}
\usepackage{graphicx}
\usepackage{adjustbox}
\usepackage{booktabs} %
\usepackage{tabularx}
\usepackage{float} %
\usepackage{color} %
\usepackage[super]{nth}
\usepackage{stmaryrd} %
\allowdisplaybreaks

\usepackage{todonotes}
\usepackage{multirow}
\usepackage{bm}

\title{\LARGE \bf
	Square-Root Inverse Filter-based GNSS-Visual-Inertial Navigation
}

\author{Jun Hu$^\dag$, Xiaoming Lang$^\dag$, Feng Zhang$^\dag$,  Yinian Mao$^\dag$, and Guoquan Huang$^\dag$$^\ddag$ %
	\thanks{$\dag$ Meituan UAV, Beijing, China (e-mail: {hujun11 $\mid$ langxiaoming $\mid$ zhangfeng53 $\mid$ maoyinian $\mid$ huangguoquan}@meituan.com).
 $\ddag$ Dept. of Mechanical Engineering, Computer and Information Sciences, University of Delaware, Newark, DE.}%
}

\begin{document}

	\newcommand{\tabincell}[2]{\begin{tabular}{@{}#1@{}}#2\end{tabular}}
	
	\maketitle

	\begin{abstract}

While Global Navigation Satellite System (GNSS) is often used to provide global positioning if available, its intermittency and/or inaccuracy  calls for fusion with other sensors.   
  In this paper, we develop a novel GNSS-Visual-Inertial Navigation System (GVINS) that fuses visual, inertial, and raw GNSS measurements within the square-root inverse sliding window filtering (SRI-SWF) framework in a tightly coupled fashion, which thus is termed SRI-GVINS. 
  In particular, for the first time, we deeply fuse the GNSS pseudorange, Doppler shift, single-differenced pseudorange, and double-differenced carrier phase measurements, along with the visual-inertial measurements. 
Inherited from the SRI-SWF, the proposed SRI-GVINS gains significant numerical stability and computational efficiency over the start-of-the-art methods. 
  Additionally, we propose to use a filter to sequentially initialize the reference frame transformation till converges, rather than collecting measurements for batch optimization.
  We also perform online calibration of GNSS-IMU extrinsic parameters to mitigate the possible extrinsic parameter degradation.
The proposed SRI-GVINS is extensively evaluated on our own collected UAV datasets and the results demonstrate that the proposed method is able to suppress VIO drift in real-time and also show the effectiveness of online GNSS-IMU extrinsic calibration. 
The experimental validation on the public datasets further reveals that the proposed SRI-GVINS outperforms the state-of-the-art methods in terms of both accuracy and efficiency.
		
	\end{abstract}

	\section{Introduction and related work}
	
	In recent years, visual-inertial odometry (VIO) or visual-inertial navigation system (VINS) has seen increasing popularity in various fields such as autonomous driving, augmented/virtual reality (AR/VR), and aerial vehicles~\cite{huang2019visual}. VIO aims to fuse camera measurements and Inertial Measurement Unit (IMU) data to estimate 3D motion of the sensing platform including the orientation, position and velocity and its algorithms can be broadly categorized into optimization-based methods \cite{qin2018vins, leutenegger2015keyframe, campos2021orb, von2022dm} and filter-based methods \cite{mourikis2007multi, li2013high, geneva2020openvins}.

        Despite its success, VIO can easily accumulate drifts and cannot provide global localization due to the four unobservable directions corresponding to the global position and yaw \cite{kelly2011visual, hesch2012observability}.  
        On the other hand, Global Navigation Satellite System (GNSS) measurements can offer absolute measurements in the global frame if available and can be leveraged to correct the VIO drift. By fusing VIO and GNSS, it is possible to achieve drift-free global localization as well as consistent local trajectories.

        There are two primary approaches for fusing VIO and GNSS: loosely-coupled and tightly-coupled. In the loosely-coupled approach, VIO and GNSS are processed independently, as demonstrated in various studies \cite{angelino2012uav, lynen2013robust, shen2014multi, schreiber2016vehicle, mascaro2018gomsf, yu2019gps, qin2019general, gong2020graph, lee2020intermittent, xiong2021g}. Some researchers \cite{angelino2012uav, lynen2013robust} have developed state estimation systems that integrate GNSS solutions with visual and inertial data using the extended Kalman filter (EKF). \cite{schreiber2016vehicle} integrated visual odometry with GNSS pseudorange measurements using an EKF framework. Other researchers \cite{mascaro2018gomsf, yu2019gps, qin2019general, gong2020graph} employed an optimization framework to fuse the results from local VIO with GNSS solutions. Lee et al. \cite{lee2020intermittent} proposed a multi-state constraint Kalman filter (MSCKF)-based estimator to optimally fuse inertial, camera, and asynchronous GNSS position and perform online calibration of the GNSS-IMU extrinsic and time offset. However, most of the aforementioned works rely on GNSS solutions for estimation and are unable to provide information in scenarios where the number of tracked satellites is less than four.

        In the tightly (or deeply)-coupled approach, VIO and GNSS are integrated within a joint optimization framework \cite{won2014gnss, li2019tight, cao2022gvins, liu2021optimization, li2022p, lee2022tightly, liu2022variable, liu2023ingvio}. This framework simultaneously optimizes GNSS raw measurements (pseudorange, carrier phase, or Doppler shift), visual, and inertial information to minimize their errors. \cite{cao2022gvins} and \cite{liu2021optimization} have proposed optimization-based approaches that jointly fuse inertial, camera feature reprojection, GNSS pseudorange, and Doppler shift measurements in a sliding window. \cite{li2022p} employed the ionosphere-free (IF) model with dual-frequency observations and incorporated phase ambiguity into the estimated states to eliminate ionospheric effects and utilize carrier phase measurements. To enhance computational efficiency, some filter-based implementations \cite{lee2022tightly, liu2022variable, liu2023ingvio} have been proposed. Lee et al. \cite{lee2022tightly}  model raw measurement uncertainties by canceling atmospheric effects.  Liu et al. \cite{liu2023ingvio} proposed an invariant filter-based platform, which derives the degenerate motions and proves intrinsic consistency. However, it should be noted that \cite{liu2023ingvio} does not support online GNSS-IMU extrinsic calibration. Additionally, these filter-based methods do not interpolate velocity at the time of Doppler shift measurement, which can introduce unmodeled errors.

        Note that both filtering-based \cite{geneva2020openvins} and optimization-based \cite{qin2018vins} VIO methods could have numerical issues \cite{maybeck1982stochastic}, primarily due to the fact that if the condition number of the covariance/Hessian matrix is larger than $10^9$, it can result in numerical errors and lead to poor estimation accuracy or even divergence. 
        In many embedded systems, single-precision arithmetic is preferred over double-precision arithmetic due to its lower hardware resource requirements and faster operations. Motivated by the superior numerical stability and potential speed gains of square-root methods \cite{bierman2006factorization}, Wu et al.~\cite{wu2015square} proposed a square-root inverse sliding window filter (SRI-SWF) for VIO. 
        Given the stringent resources available on unmanned aerial vehicles (UAVs) which  we are interested in this work, the SRI-SWF becomes a promising solution.
        
        In this paper, we propose a tightly-coupled GNSS-Visual-Inertial Navigation System (GVINS) that {\em deeply} fuses GNSS raw measurements within an efficient square-root filtering framework.
        In particular, our main contributions include:
	\begin{itemize}
		
		\item For the first time, we deeply fuse the raw GNSS data including pseudorange, Doppler shift, single-differenced pseudorange and double-differenced carrier phase, with IMU and camera measurements, within the SRI-SWF framework, in order to provide drift-free global localization. The proposed SRI-GVINS offers compelling numerical stability and computational efficiency.
		
  \item We develop a filter-based initialization approach to sequentially and adaptively converge the transformation between the reference frame and the VIO frame. Furthermore, we leverage IMU integration to update the state using delayed and asynchronous GNSS measurements. Additionally, we perform online GNSS-IMU extrinsic calibration to mitigate the adverse effects of extrinsic parameter degradation.
		\item We validate the proposed SRI-GVINS on UAV real-world datasets, demonstrating its ability to eliminate VIO drift in real-time and the effectiveness of online GNSS-IMU extrinsic calibration. Moreover, the experimental results on the public datasets reveal that the proposed SRI-GVINS achieves superior accuracy and efficiency compared to the state-of-the-art methods such as GVINS \cite{cao2022gvins} and InGVIO \cite{liu2023ingvio}.
		
	\end{itemize}

	\section{Square-Root Inverse Sliding-Window Filter}
	
\begin{figure}%
		\centering
		\includegraphics[scale=0.25]{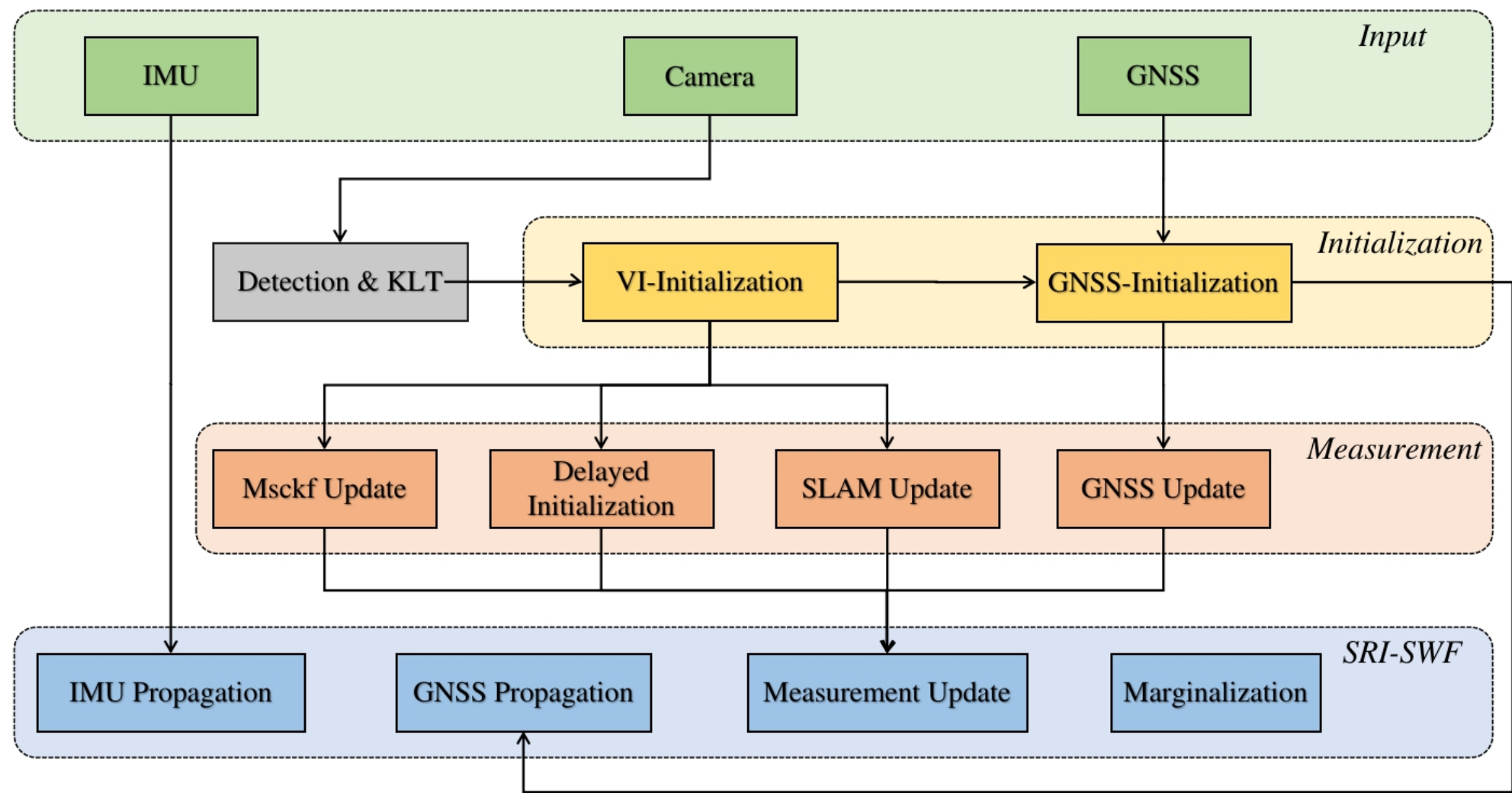}
		\caption{Overview of the proposed SRI-GVINS system.}
		\label{framework}
	\end{figure}

In this section, we briefly review the SRI-SWF~\cite{wu2015square}, 
which serves as the foundation for the proposed SRI-GVINS
whose architecture is  shown in Fig. \ref{framework}.

	At each time step $k$, the state vector $\textbf{x}_k$ comprises the current SLAM features $\textbf{x}_S$, historical IMU pose clones $\textbf{x}_{C}$, parameter state vector $\textbf{x}_P$, and IMU extra state $\textbf{x}_E$.
	\begin{align}
	\textbf{x}_k &=[\textbf{x}_S^\top \ \ \textbf{x}_C^\top \ \   \textbf{x}_P^\top \ \ \textbf{x}_E^\top]^\top, \quad \textbf{x}_S =[\textbf{p}_{f_1}^\top \ \dots \ \textbf{p}_{f_n}^\top]^\top\\
        \textbf{x}_{C} &=[\textbf{x}_{C_{k-M+1}}^\top \ \dots \ \textbf{x}_{C_k}^\top]^\top, \quad \textbf{x}_{C_i} =[{}^W\textbf{q}_{I_i}^\top \ \ {}^W\textbf{p}_{I_i}^\top]^\top \label{x_C}\\
	\textbf{x}_P &=[^C\textbf{q}_I^\top \ \ ^C\textbf{p}_I^\top   \ \ ^I\textbf{p}_r^\top \ \ \mathbf {\delta t}_{0}^{r}  \ \dots \  \mathbf {\delta t}_{N-1}^{r}\ \ \mathbf {\delta \dot{t} }^{r}]^\top \label{x_P} \\
        \textbf{x}_E &=[^W\textbf{v}_{I_k}^\top  \ \   \textbf{b}_{a_k}^\top  \ \   \textbf{b}_{g_k}^\top]^\top \label{x_E}
 \end{align}
 where $\textbf{p}_{f_j}$ for $j = 1,\dots, n$ donotes the state of point feature $\textbf f_j$, represented by the inverse-depth parameterization \cite{civera2008inverse} with respect to its first observing camera pose within the current sliding window. $\textbf{x}_{C_i}$ represents the state of cloned IMU pose at time step $i$, where $^W\textbf{q}_{I_i}$ is the quaternion describing the rotation from the IMU frame $\left \{ I_i \right \} $ to the local VIO world frame $\left \{ W \right \} $, and $^W\textbf{p}_{I_i}$ is the position of $\left \{ I_i \right \} $ in $\left \{ W \right \} $. $[^C\textbf{q}_I \ ^C\textbf{p}_I]$ is the IMU to camera extrinsic parameters, $^I\textbf{p}_r$ is the GNSS receiver to IMU extrinsic parameters. $\mathbf {\delta t}_{j}^{r}$ for $j = 0,\dots, N-1$ donotes the receiver clock bias with respect to GNSS constellation $j$. $\mathbf {\delta \dot{t} }^{r}$ is the receiver clock drift rate, which is the same for each constellation. $^W\textbf{v}_{I_k}$ is the velocity of $\left \{ I \right \} $ in $\left \{ W \right \} $, while $\textbf{b}_{a_k}$ and $\textbf{b}_{g_k}$ are the accelerometer and gyroscope biases at time step $k$.

	\subsection{IMU Propagation}
	Given the inertial measurement $\mathbf{u}_{k,k+1}=\begin{bmatrix}\boldsymbol{\omega}_{m_{k}}^{T}&\textbf{a}_{m_{k}}^{T}\end{bmatrix}^{T}$, we utilize the inertial kinematics \cite{trawny2005indirect} to propagate as:
     \begin{align}
     \mathbf{x}_{I_{k+1}}=\mathbf{f}(\mathbf{x}_{I_{k}},\mathbf{u}_{k,k+1}-\mathbf{w}_{k,k+1}) \label{prop}
     \end{align}
     where $\mathbf{x}_{I_{k}} = \begin{bmatrix}\mathbf{x}_{C_{k}}^{T}&\mathbf{x}_{E_{k}}^{T}\end{bmatrix}^{T}$, and $\mathbf{w}_{k,k+1}$ is the Gaussian noise affecting the IMU measurements with covariance $\mathbf{Q}_k$.

     By linearizing Eq. (\ref{prop}) around the state estimates, we obtain the error-state propagation equation:
     \begin{align}
{\tilde{\mathbf{x}}}_{I_{k+1}}=\boldsymbol{\Phi}_{k+1,k}\widetilde{\mathbf{x}}_{I_{k}}+\mathbf{G}_{k+1,k}\mathbf{w}_{k,k+1}
    \end{align}
where $\boldsymbol{\Phi}_{k+1,k}$ and $\mathbf{G}_{k+1,k}$ are the corresponding Jacobians. 
    
    In the IMU propagation step, a new pose state $\mathbf{x}_{I_{k+1}}$ [see Eq. (\ref{prop})] is added to the current state vector:
    \begin{align}
    \mathbf{x}_{k+1}^{\ominus}=\begin{bmatrix}\mathbf{x}_{k}^T&\mathbf{x}_{I_{k+1}}^T\end{bmatrix}^T
    \end{align}

    The prior (upper-triangular) square-root information matrix and residual vector are defined as follows:
    \begin{align}
    \begin{split}
    &\mathbf{R}_{k+1}^{\ominus} =\begin{bmatrix}\mathbf{R}_{k}&\mathbf{0}\\\mathbf{V}_{1}&-\mathbf{Q}_{k}^{\prime{-\frac{1}{2}}}\end{bmatrix},\quad\mathbf{r}_{k+1}^{\ominus} = \mathbf{0}  \\
    &\mathbf{V}_{1} =\begin{bmatrix}\mathbf{0}\cdots\mathbf{0}&\mathbf{Q}_{k}^{'-\frac{1}{2}}\mathbf{\Phi}_{k+1,k}^{(C)}&\mathbf{0}\cdots\mathbf{0}&\mathbf{Q}_{k}^{'-\frac{1}{2}}\mathbf{\Phi}_{k+1,k}^{(E)}\end{bmatrix} \label{prop_V1}
    \end{split}
    \end{align}
where $\mathbf{Q}_{k}^{\prime}=\mathbf{G}_{k+1,k}\mathbf{Q}_{k}\mathbf{G}_{k+1,k}^{T}$, $\mathbf{\Phi}_{k+1,k}^{(C)}$ and $\mathbf{\Phi}_{k+1,k}^{(E)}$ are the block columns of the Jacobian $\mathbf{\Phi}_{k+1,k}$ with respect to the clone and extra IMU states, respectively. Note that $\mathbf{R}_{k+1}^{\ominus}$ is not upper-triangular before the next marginalization step.
	
	\subsection{Marginalization}
	At each time step $k$, the past SLAM features $\widetilde{\mathbf{x}}_{DS}$, the oldest clone $\widetilde{\mathbf{x}}_{C_{k-M}}$, and the extra IMU state $\widetilde{\mathbf{x}}_{E_{k-1}}$ are marginalized. The (error) state vector that includes all these states to be marginalized is defined as follows:
        \begin{align}
        \widetilde{\mathbf{x}}_k^M=\begin{bmatrix}\widetilde{\mathbf{x}}_{DS}^T&\widetilde{\mathbf{x}}_{C_{k-M}}^T&\widetilde{\mathbf{x}}_{E_{k-1}}^T\end{bmatrix}^T
	\end{align}

        We donate $\widetilde{\mathbf{x}}_k^R$ as the remaining states, and through a permutation matrix, the state vector is transformed as follows:
        \begin{align}
        \mathbf{P}_M\widetilde{\mathbf{x}}_k^{\ominus}=\left[\widetilde{\mathbf{x}}_k^{M^T}\quad\widetilde{\mathbf{x}}_k^{R^T}\right]^T
        \end{align}
        
        After performing a full QR decomposition, as described in \cite{wu2015square}, we obtain the prior(upper-triangular) square-root information matrix and residual vector, with only $\widetilde{\mathbf{x}}_k^{R}$ retained.
	
        \subsection{Measurement Update}
	
	In general, a measurenment can be modeled as follows:
        \begin{align}
        \mathbf{z}=\mathbf{h}(\mathbf{x})+\mathbf{n},\quad\text{with n}\sim\mathcal{N}(\mathbf{0},\mathbf{\Sigma}) \label{update}
        \end{align}
where $\mathbf{h}$ is a nonlinear function that describes the measurement
model, $\mathbf{z}$ is the obtained measurement(e.g., SLAM feature reobservations and MSCKF feature measurements), $\mathbf{n}$ and $\mathbf{\Sigma}$ represent the white Gaussian noise of the measurement and its covariance, respectively.

        By linearizing Eq. (\ref{update}) around the state estimate $\mathbf{\hat{x}}$, we obtain the linearized measurement constraint $\mathbf{r}\simeq\mathbf{H}\widetilde{\mathbf{x}}+\mathbf{n}$, 
 where $\mathbf{H}=\frac{\partial\mathbf{h}}{\partial\mathbf{x}}\bigg|_{\mathbf{x}=\hat{\mathbf{x}}}$ and $\mathbf{r}=\mathbf{z}-\mathbf{h}(\hat{\mathbf{x}})$ represent the measurement Jacobian matrix and residual, respectively. Finally, we can update the state by performing a QR decomposition:
\begin{align}
&\begin{bmatrix}\mathbf{R}&\mathbf{0}\\\mathbf{\Sigma}^{-\frac{1}{2}}\mathbf{H}&\mathbf{\Sigma}^{-\frac{1}{2}}\mathbf{r}\end{bmatrix}\stackrel{\mathrm{QR}}{=}\mathbf{Q}\begin{bmatrix}\mathbf{R}^{\oplus}&\mathbf{r}^{\oplus}\\\mathbf{0}&\mathbf{e}\end{bmatrix} \\
&\mathbf{R}\leftarrow\mathbf{R}^{\oplus}, \
\Delta\mathbf{x}=\mathbf{R}^{\oplus^{-1}}\mathbf{r}^{\oplus}
\end{align}

 \section{SRI-GVINS: Initialization and Update}

  Within the SRI-SWF framework, as shown in Fig. \ref{framework}, 
  In addition to the visual-inertial measurements as in VIO,
  the proposed SRI-GVINS deeply fuses the GNSS measurements of pseudorange, Doppler shift, single-differenced pseudorange and double-differenced carrier phase.
  Therefore, we here focus on the measurement update with raw GNSS data.

	\subsection{Frame Initialization}
    
 The GNSS update requires the transformation from the local VIO World frame to the ECEF frame $\left \{ ^E\textbf{R}_W \ ^E\textbf{p}_W \right \} $, the receiver clock bias $[\mathbf {\delta t}_{0}^{r}  \ \dots \  \mathbf {\delta t}_{N-1}^{r}]$, and the  receiver clock drift rate $\mathbf {\delta \dot{t} }^{r}$ [see (\ref{x_P})].
	To initialize the transform $\left \{ ^E\textbf{R}_W \ ^E\textbf{p}_W \right \} $, we insert the transform variables into the filter to adaptively converge the 4 degrees-of-freedom (DOF) transform. This approach differs from  \cite{lee2020intermittent}, which collects measurements of a desired trajectory length to solve a least squares estimation problem. The position and velocity of the receiver in VIO world frame can be expressed as:
        \begin{align}
        &{}^{W}\textbf{p}_{r}={}^{W}\textbf{p}_{I}+{}^W\textbf{R}_I {}^I\textbf{p}_r \label{pos_w_r} \\
        &{}^{W}\textbf{v}_{r}={}^{W}\textbf{v}_{I}+{}^W\textbf{R}_I \lfloor{}^{I}\omega_{I}\rfloor{}^I\textbf{p}_r \label{vel_w_r}
        \end{align}
        
        By utilizing the first GNSS measurement as the reference point, we employ GNSS Single Point Positioning(SPP) output to obtain the receiver's position and velocity in the North-East-Down frame (NED, $\left \{ N \right \} $). At each time step $k$, we have the following geometric constraints:
        \begin{align}
        & {}^{N}\textbf{p}_{r_k}={}^{N}\textbf{p}_{W}+{}^N\textbf{R}_{W} {}^W\textbf{p}_{r_k}, \ {}^{N}\textbf{v}_{r_k}={}^N\textbf{R}_{W} {}^W\textbf{v}_{r_k} \label{N_p_rk}
        \end{align}

        Note that the receiver's position and velocity at time step $k$ can be predicted from the GNSS measurement at time step $i$ through IMU integration. Since ${}^N\textbf{R}_{W}$ is 1 DOF for the yaw angle, the transform states can be represented as $[\theta \ \  ^N\textbf{p}_W]$.

        By augmenting the state vector with the transform variables along with an infinite covariance prior, we perform the standard measurement update [see Eq. (\ref{N_p_rk})]. We determine that the transform has converged when the covariance matrix $P_{\theta \theta}$ for the yaw angle falls below a decision threshold of $\sigma_{\theta}^{2}$. After initialization, we marginalize the transform states. Given $\left \{ ^N\textbf{R}_W \ ^N\textbf{p}_W \right \} $ and $\left \{ ^E\textbf{R}_N \ ^E\textbf{p}_N \right \} $ computed from the reference point, the transform $\left \{ ^E\textbf{R}_W \ ^E\textbf{p}_W \right \} $ is obtained as:
        \begin{align}
        & {}^E\textbf{R}_{W} = {}^E\textbf{R}_{N}{}^N\textbf{R}_{W}, \ {}^{E}\textbf{p}_{W}={}^{E}\textbf{p}_{N}+{}^E\textbf{R}_N {}^N\textbf{p}_W
        \end{align}

	\subsection{GNSS Receiver Clock Bias}
	
	We use the pseudorange  and Doppler shift measurements to initialize the GNSS receiver clock bias $[\mathbf {\delta t}_{0}^{r}  \ \dots \  \mathbf {\delta t}_{N-1}^{r}]$ and the receiver clock drift rate $\mathbf {\delta \dot{t} }^{r}$.
 To this end,  we calculate the residual of the pseudorange measurement ${}^{s_{j}}{p}_{r_{k}}$ at time step $k$ as follows:
        \begin{align}
        \begin{split}
        \mathbf{z_p}_{r_{k}}^{s_{j}}= &\|{}^E\textbf{p}_{s_j}-{}^E\textbf{R}_{W}{}^W\textbf{p}_{r_k}-{}^{E}\textbf{p}_{W}\|-{}^{s_{j}}{p}_{r_{k}} \\
        & + c(\mathbf {\delta t}_{c_j}^{r}-\Delta t^{s_{j}})+T_{r_{k}}^{s_{j}}+I_{r_{k}}^{s_{j}}
        +S_{r_{k}}^{s_{j}} + n_{pr} \label{zp}
        \end{split}
        \end{align}
        $c\approx 3.0\times 10^{8}m/s$ denotes the speed of light. $\mathbf {\delta t}_{c_j}^{r}$ and $\Delta t^{s_{j}}$ refer to the clock bias of the receiver and satellite $s_{j}$, respectively. $T_{r_{k}}^{s_{j}}$, $I_{r_{k}}^{s_{j}}$ and $S_{r_{k}}^{s_{j}}$ represent the tropospheric delay, ionospheric delay and Sagnac term, respectively. $n_{pr}$ represents the pseudorange measurement noise. 
        
        Similarly, the residual of the Doppler shift measurement ${}^{s_{j}}{d}_{r_{k}}$ is given by:
        \begin{align}
        \begin{split}
        \mathbf{z_d}_{r_{k}}^{s_{j}}= &\frac{1}{\lambda}\mathbf{\kappa}_{r_{k}}^{s_{j}}{}^{T}({}^E\textbf{v}_{s_j}-{}^E\textbf{R}_{W}{}^W\textbf{v}_{r_k})
        + \frac{c}{\lambda}(\mathbf {\delta \dot{t} }^{r}-\Delta\dot{t}^{s_{j}}) \\
        &+{}^{s_{j}}{d}_{r_{k}} + n_d \label{zd}
        \end{split}
        \end{align}
where $\lambda$ represents the GNSS carrier wavelength. $\mathbf{\kappa}_{r_{k}}^{s_{j}}$ is the unit direction vector from the receiver to satellite in $\left \{ E \right \} $. $^E\textbf{v}_{s_j}$ denotes the velocity of satellite $s_{j}$ in $\left \{ E \right \} $. $\mathbf {\delta \dot{t} }^{r}$ and $\Delta\dot{t}^{s_{j}}$ refer to the clock drift rate of the receiver and satellite $s_{j}$, respectively. 
$n_d$ represents the measurement noise.

        By using all the GNSS measurement residuals [see Eq. (\ref{zp}) and Eq. (\ref{zd})], we solve the following linear constraint $\mathbf{A}\mathbf{X}=\mathbf{b}$, where $\mathbf{X}$ is a stack of $[\mathbf {\delta t}_{0}^{r}  \ \dots \  \mathbf {\delta t}_{N-1}^{r}]$ and $\mathbf {\delta \dot{t} }^{r}$. Initially, we apply the RANSAC algorithm \cite{fischler1981random} to eliminate any erroneous constraints. Subsequently, we solve the least squares problem  mentioned above to obtain the initial guesses of the GNSS receiver clock bias and clock drift rate. Afterward, we further augment the initial values into the state through delayed initialization \cite{li2014visual}.

        Once initialized, we propagate $[\mathbf {\delta t}_{0}^{r}  \ \dots \  \mathbf {\delta t}_{N-1}^{r}]$ and $\mathbf {\delta \dot{t} }^{r}$ as follows:
        \begin{align}
        & \mathbf {\delta t}_{k}^{r}=\mathbf {\delta t}_{k-1}^{r}+\mathbf{1}_{N\times1}\mathbf {\delta \dot{t} }^{r}(t_{k} - t_{k-1})
        , \ \mathbf {\delta \dot{t}}^{r}_{t_k} = \delta \mathbf {\dot{t}}^{r}_{t_{k-1}}
        \end{align}
        The prior (upper-triangular) square-root information matrix [see Eq. (\ref{prop_V1})] is given by:
        \begin{align}
        \begin{split}
        &\mathbf{R}_{k+1}^{\ominus} =\begin{bmatrix}\mathbf{R}_{k}&\mathbf{0}\\\mathbf{V}_{1}&-\mathbf{Q}_{k}^{\prime{-\frac{1}{2}}}\end{bmatrix},\quad\mathbf{r}_{k+1}^{\ominus} = \mathbf{0}  \\
        &\mathbf{V}_{1} =\begin{bmatrix}\mathbf{0}\cdots\mathbf{0}&\mathbf{Q}_{k}^{'-\frac{1}{2}}\mathbf{\Phi}_{k+1,k}^{(C)}&\mathbf{Q}_{k}^{'-\frac{1}{2}}\mathbf{\Phi}_{k+1,k}^{(P)}&\mathbf{Q}_{k}^{'-\frac{1}{2}}\mathbf{\Phi}_{k+1,k}^{(E)}\end{bmatrix} 
        \end{split}
        \end{align}
        where $\mathbf{\Phi}_{k+1,k}^{(P)}$ represents the block columns of the Jacobian $\mathbf{\Phi}_{k+1,k}$ with respect to the parameter state.

 \subsection{GNSS Measurement Update}
 
	Due to the delay and asynchronous arrival of GNSS messages, the GNSS measurement time is typically earlier than the inertial state time. Since the velocity of each cloned pose is not preserved in the sliding window, it is not feasible to obtain the velocity state of the GNSS measurement time through interpolation\cite{lee2022tightly}. Assuming the inertial state at time step $i$ and the GNSS measurement at time step $k$, the following constraints hold through IMU integration:
        \begin{align}
        \begin{split}
        &{}^W\textbf{R}_{I_k}={}^W\textbf{R}_{I_i}(\mathbf\gamma_{I_i}^{I_k})^{T}\\
        &{}^W\textbf{v}_{I_k}={}^W\textbf{v}_{I_i} - {}^W\textbf{g}{\Delta t} - {}^W\textbf{R}_{I_k}\mathbf\beta_{I_i}^{I_k}\\
        &{}^W\textbf{p}_{I_k}={}^W\textbf{p}_{I_i}-{}^W\textbf{v}_{I_k}{\Delta t}-\frac{1}{2}{}^W\textbf{g}{\Delta t}^{2}-{}^W\textbf{R}_{I_k}\mathbf\alpha_{I_i}^{I_k} \label{integrate_meas}
        \end{split}
        \end{align}
where $\mathbf\gamma_{I_i}^{I_k}$, $\mathbf\beta_{I_i}^{I_k}$ and $\mathbf\alpha_{I_i}^{I_k}$ are the pre-integration terms~\cite{forster2016manifold}. For inertial measurements within the time interval $[t_k, t_i]$, these terms are computed as follows:
        \begin{align}
        \begin{split}
        & \mathbf{\gamma}_{I_i}^{I_k} =\prod_{j=k}^{i-1}\operatorname{Exp}\left((\boldsymbol{\omega}_{j}-\mathbf{b}_{w_{j}})\Delta t\right)\\
        &\mathbf\beta_{I_i}^{I_k} =\sum_{j=k}^{i-1}\gamma_{I_j}^{I_k}(\mathbf{a}_{j}-\mathbf{b}_{a_{j}})\Delta t \\
        & \mathbf\alpha_{I_i}^{I_k} =\sum_{j=k}^{i-1}\left [ \beta_{I_j}^{I_k}\Delta t+\frac{1}{2} \gamma_{I_j}^{I_k}(\mathbf{a}_{j}-\mathbf{b}_{a_{j}}){\Delta t}^{2}\right ]
        \end{split}
        \end{align}

        \subsubsection{Pseudorange Measurement (Psr)}

        We substitute Eq. (\ref{pos_w_r}) and Eq. (\ref{integrate_meas}) into Eq. (\ref{zp}) to define the complete residual of the pseudorange measurement.

        \subsubsection{Doppler shift Measurement (Dopp)}
        We substitute Eq. (\ref{vel_w_r}) and Eq. (\ref{integrate_meas}) into Eq. (\ref{zd}) to define the complete residual of the Doppler shift measurement.

        \subsubsection{Single-differenced Pseudorange Measurement (Dpsr)}
        Considering the presence of a base station, we can define the pseudorange measurement of the base station at time step $k$:
        \begin{align}
        \begin{split}
        \mathbf{z_p}_{b_{k}}^{s_{j}}= &\|{}^E\textbf{p}_{s_j}-{}^E\textbf{p}_{b_k}\|-{}^{s_{j}}{p}_{b_{k}} + c(\mathbf {\delta t}_{c_j}^{b} -\Delta t_b^{s_{j}})\\
        & +T_{b_{k}}^{s_{j}}+I_{b_{k}}^{s_{j}}
        +S_{b_{k}}^{s_{j}} + n_{pb} \label{zp_b}
        \end{split}
        \end{align}

        When the base station and receiver are relatively close, the ionospheric and tropospheric delays should be approximately canceled by subtraction of Eq. [\ref{zp}] and Eq. [\ref{zp_b}]:
        \begin{align}
        \begin{split}
        & \mathbf{z_{dp}}_{r_{k}}^{s_{j}}= \|{}^E\textbf{p}_{s_j}-{}^E\textbf{R}_{W}{}^W\textbf{p}_{r_k}-{}^{E}\textbf{p}_{W}\| - \|{}^E\textbf{p}_{s_j}-{}^E\textbf{p}_{b_k}\| \\
        & \ \ -{}^{s_{j}}{p}_{r_{k}} +{}^{s_{j}}{p}_{b_{k}} + c(\mathbf {\delta t}_{c_j}^{r}-\Delta t^{s_{j}} -\mathbf {\delta t}_{c_j}^{b} +\Delta t_b^{s_{j}})\\
        & \ \ +T_{r_{k}}^{s_{j}}-T_{b_{k}}^{s_{j}}+I_{r_{k}}^{s_{j}}-I_{b_{k}}^{s_{j}}
        +S_{r_{k}}^{s_{j}}-S_{b_{k}}^{s_{j}} + n_{pr} - n_{pb} \label{z_dp}
        \end{split}
        \end{align}

        \subsubsection{Double-differenced Carrier Phase Measurement (Ddcp)}
        The carrier phase measurements are often more precise than pseudorange measurements, the residual of the carrier phase measurement is given by: 
        \begin{align}
        \begin{split}
        & \mathbf{z_{cp}}_{r_{k}}^{s_{j}}= \|{}^E\textbf{p}_{s_j}-{}^E\textbf{R}_{W}{}^W\textbf{p}_{r_k}-{}^{E}\textbf{p}_{W}\|+{\lambda}N \\
        & \ \ \ - {}^{s_{j}}{cp}_{r_{k}} + c(\mathbf {\delta t}_{c_j}^{r}-\Delta t^{s_{j}})+T_{r_{k}}^{s_{j}}-I_{r_{k}}^{s_{j}} + n_{cp} \label{z_cp}
        \end{split}
        \end{align}
        where $N$ is the integer ambiguity, ${}^{s_j}{cp}_{r_k}$ is the carrier phase measurement and $n_{cp}$ is its noise. By using differential measurements between base station and base satellite $s_i$, the residual of the double-differenced carrier phase measurement can be defined as:
        \begin{align}
        \begin{split}
        & \mathbf{z}_{ddcp}^{k}= d_{s_j,r_k} - d_{s_j,b_k} - d_{s_i,r_k} + d_{s_i,b_k} +{\lambda}N_{dd} \\
        & \ \ \ -({}^{s_j}{cp}_{r_k} - {}^{s_j}{cp}_{b_k} - {}^{s_i}{cp}_{r_k} + {}^{s_i}{cp}_{b_k}) + n_{ddcp} \label{z_ddcp}
        \end{split}
        \end{align}
        where $d_{s_j,r_k}=\|{}^E\textbf{p}_{s_j}-{}^E\textbf{p}_{r_k}\|$ and $N_{dd}$ is the double-differenced integer ambiguity acquried from RTK algorithm.
	
        Jacobians of these GNSS measurements can be computed:
        \begin{align}
        & \frac{\partial \tilde{\mathbf{z}}}{\partial {}^W\tilde{\mathbf{X}}_{I_i}}=\frac{\partial \tilde{\mathbf{z}}}{\partial {}^W\tilde{\mathbf{X}}_{r_k}}\frac{\partial {}^W\tilde{\mathbf{X}}_{r_k}}{\partial {}^W\tilde{\mathbf{X}}_{I_k}}\frac{\partial {}^W\tilde{\mathbf{X}}_{I_k}}{\partial {}^W\tilde{\mathbf{X}}_{I_i}}
        \end{align}
        The overall measurement covariance matrix $\mathbf{\Sigma}$ is given by:
        \begin{align}
        \mathbf{\Sigma} = \mathbf{\Sigma}_{m} + \mathbf{J} \mathbf{\Sigma}_{g} \mathbf{J}^T, \mathbf{J} = \frac{\partial \tilde{\mathbf{z}}}{\partial {}^W\tilde{\mathbf{X}}_{r_k}}\frac{\partial {}^W\tilde{\mathbf{X}}_{r_k}}{\partial {}^W\tilde{\mathbf{X}}_{I_k}}
        \end{align}
        where $\mathbf{\Sigma}_{m}$ is the GNSS raw measurement covariance matrix, and $\mathbf{\Sigma}_{g}$ is the noise from IMU integration mesurements.

	\subsection{Online GNSS-IMU Calibration}
	
	The GNSS-IMU extrinsic is represented as $^I\textbf{p}_r$ in the state vector [see Eq. (\ref{x_P})]. We perform extrinsic estimation using the Doppler shift measurement and only update the extrinsic when the IMU excitation is sufficient. The Jacobian of the Doppler shift residual with respect to the extrinsic is:
	\begin{align}
        \frac{\partial \tilde{\mathbf{z_d}}_{r_{k}}^{s_{j}}}{\partial{}^I\tilde{\textbf{p}}_{r}}=\frac{\partial \tilde{\mathbf{z_d}}_{r_{k}}^{s_{j}}}{\partial {}^W\tilde{\textbf{p}}_{r_k}}\frac{\partial {}^W\tilde{\textbf{p}}_{r_k}}{\partial {}^I\tilde{\textbf{p}}_{r}} + \frac{\partial \tilde{\mathbf{z_d}}_{r_{k}}^{s_{j}}}{\partial {}^W\tilde{\textbf{v}}_{r_k}}\frac{\partial {}^W\tilde{\textbf{v}}_{r_k}}{\partial {}^I\tilde{\textbf{p}}_{r}}
        \end{align}

	\section{EXPERIMENTS ON UAV DATASETS}
	We first validate the proposed SRI-GVINS on our own collected UAV real-world datasets, the groundtruth is provided by an RTK-INS. The sensor platform consisted of two downward-looking cameras running at 20Hz with resolution 1920$\times$1080, an IMU running at 250Hz and a Unicore UM982 GNSS receiver running at 5Hz. The real-time RTCM stream is broadcasted from a self-built base station located 3km away, which enables us to use single-differenced pseudorange and double-differenced carrier phase measurements.

	\subsection{GNSS Measurements}
\begin{figure}[thpb]
		\centering
		\includegraphics[scale=0.25]{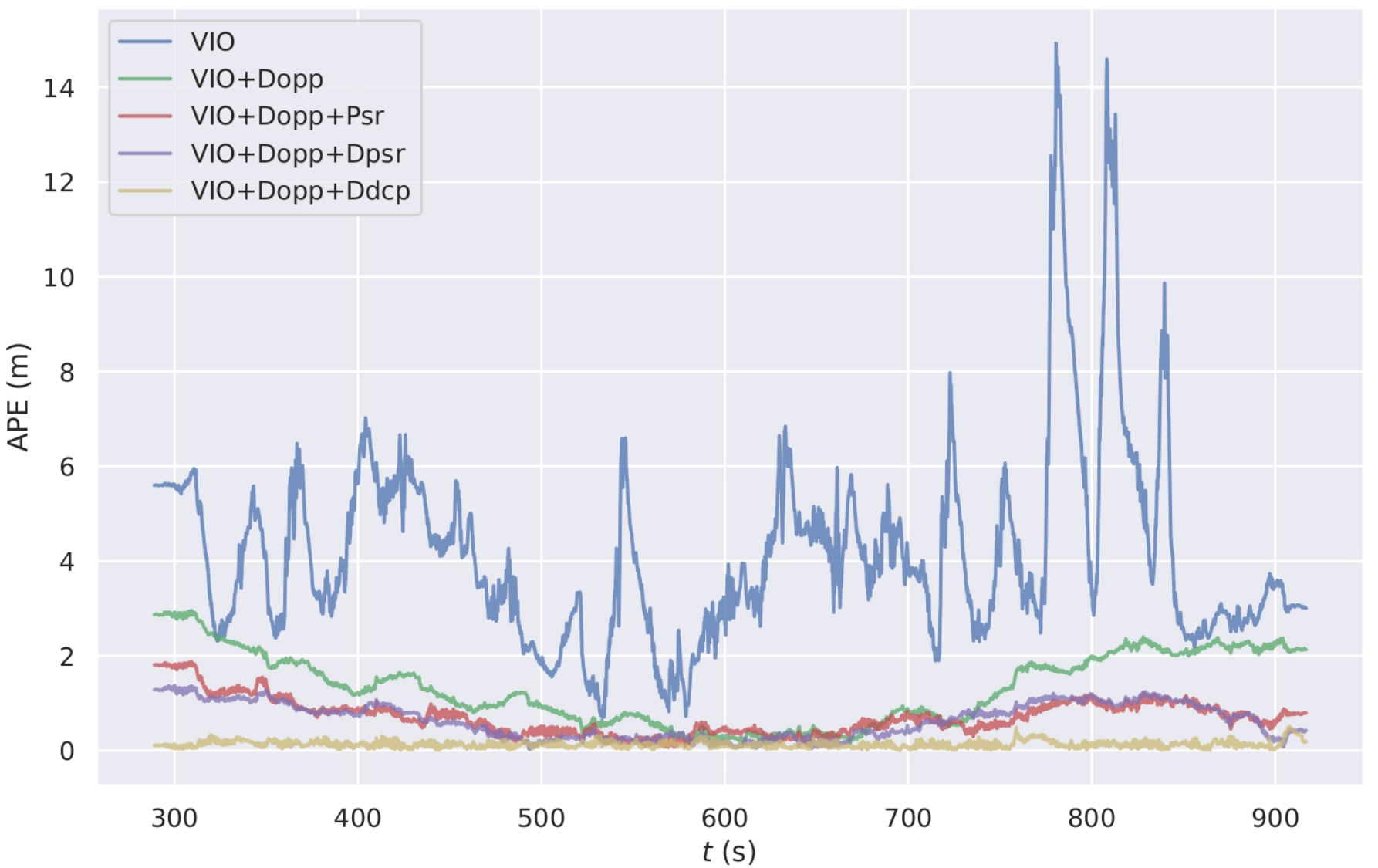}
		\caption{Position errors of SRI-GVINS using different GNSS measurements in the 50m-3km-8m/s dataset.}
		\label{v4_evo}
	\end{figure}

	To investigate the effect of GNSS measurement update, we evaluated the performance of SRI-GVINS using different GNSS measurements. We selected 10 outdoor UAV datasets with varying heights ranging from 20m to 60m, trajectory lengths ranging from 400m to 3km, and flying speeds ranging from 4m/s to 12m/s. 
 Fig. \ref{v4_evo} shows the position errors of the {50m-3km-8m/s} dataset.

        \subsubsection{Psr and Dopp Measurements}
        Since Psr [see Eq. (\ref{zp})] and Dopp [see Eq. (\ref{zd})] measurements do not require a base station, we conducted tests using the following configurations in the SRI-GVINS system: VIO no-fusion with GNSS measurements(VIO), VIO loosely-coupled fusion with GPS velocity(VIO+Gps\_vel), VIO tightly-coupled fusion with Dopp(VIO+Dopp), and VIO tightly-coupled fusion with Dopp and Psr(VIO+Dopp+Psr). The root mean squared error (RMSE) \cite{sturm2012benchmark} for translation and body velocity estimates on each dataset are shown in Table \ref{psr_dopp_table}. 

        From Table \ref{psr_dopp_table}, we can see that the VIO+Dopp+Psr method exhibited the highest average position and velocity accuracy, demonstrating its ability to eliminate VIO drift. It is also observed that the VIO+Dopp method has higher accuracy than the VIO+Gps\_vel method. Additionally, it is interesting to note that fusing Psr on top of the VIO+Dopp fusion can improve position accuracy on long trajectory datasets, but may result in a slight decrease in position accuracy on short trajectory datasets.

	\begin{table}[]
		\caption{RMSE with Psr and Dopp measurements}
		\label{psr_dopp_table}
		\begin{center}
			\begin{tabular}{c c c c c}
				\hline
				\tabincell{l}{Height-Len \\ -Vel} & \tabincell{l}{VIO \\ +Dopp+Psr} & \tabincell{l}{VIO \\ +Dopp} & \tabincell{l}{VIO \\ +Gps\_vel} & VIO\\
				\hline
				\tabincell{l}{60m-1km \\ -6m/s} & \tabincell{l}{Tran. 0.652 \\ Vel. 0.0813} & \tabincell{l}{Tran. \textbf{0.440} \\ Vel. \textbf{0.0767}} & \tabincell{l}{Tran. 0.656 \\ Vel. 0.0978} & \tabincell{l}{Tran. 2.915 \\ Vel. 0.2170}\\
				\hline
				\tabincell{l}{60m-1km \\ -4m/s} & \tabincell{l}{Tran. 0.509 \\ Vel. \textbf{0.0706}}  & \tabincell{l}{Tran. \textbf{0.492} \\ Vel. 0.0709} & \tabincell{l}{Tran. 0.707 \\ Vel. 0.0952} & \tabincell{l}{Tran. 1.758 \\ Vel. 0.1842}\\
				\hline
				\tabincell{l}{60m-1km \\ -8m/s} & \tabincell{l}{Tran. 0.522 \\ Vel. \textbf{0.0933}}  & \tabincell{l}{Tran. \textbf{0.410} \\ Vel. 0.0938} & \tabincell{l}{Tran. 0.595 \\ Vel. 0.1043} & \tabincell{l}{Tran. 4.641 \\ Vel. 0.3361}\\
				\hline
				\tabincell{l}{60m-1km \\ -10m/s} & \tabincell{l}{Tran. 0.642 \\ Vel. 0.0936}  & \tabincell{l}{Tran. \textbf{0.4981} \\ Vel. \textbf{0.0922}} & \tabincell{l}{Tran. 0.854 \\ Vel. 0.0983} & \tabincell{l}{Tran. 4.176 \\ Vel. 0.3134}\\
				\hline
				\tabincell{l}{60m-400m \\ -8m/s} & \tabincell{l}{Tran. 0.436 \\ Vel. 0.0690}  & \tabincell{l}{Tran. \textbf{0.276} \\ Vel. \textbf{0.0684}} & \tabincell{l}{Tran. 0.394 \\ Vel. 0.0769} & \tabincell{l}{Tran. 2.628 \\ Vel. 0.2527}\\
				\hline
				\tabincell{l}{50m-3km \\ -4m/s} & \tabincell{l}{Tran. \textbf{0.707} \\ Vel. \textbf{0.0756}}  & \tabincell{l}{Tran. 1.434 \\ Vel. 0.0774} & \tabincell{l}{Tran. 0.902 \\ Vel. 0.0868} & \tabincell{l}{Tran. 16.090 \\ Vel. 0.2698}\\
				\hline
				\tabincell{l}{50m-3km \\ -8m/s} & \tabincell{l}{Tran. \textbf{0.804} \\ Vel. 0.0892}  & \tabincell{l}{Tran. 1.522 \\ Vel. \textbf{0.0869}} & \tabincell{l}{Tran. 1.398 \\ Vel. 0.0990} & \tabincell{l}{Tran. 4.634 \\ Vel. 0.3017}\\
				\hline
				\tabincell{l}{50m-3km \\ -12m/s} & \tabincell{l}{Tran. \textbf{0.694} \\ Vel. 0.1006}  & \tabincell{l}{Tran. 0.992 \\ Vel. \textbf{0.0987}} & \tabincell{l}{Tran. 1.171 \\ Vel. 0.1207} & \tabincell{l}{Tran. 3.988 \\ Vel. 0.3056}\\
				\hline
				\tabincell{l}{20m-1km \\ -8m/s} & \tabincell{l}{Tran. 0.467 \\ Vel. \textbf{0.1267}}  & \tabincell{l}{Tran. \textbf{0.381} \\ Vel. 0.1401} & \tabincell{l}{Tran. 0.673 \\ Vel. 0.1378} & \tabincell{l}{Tran. 8.614 \\ Vel. 0.7148}\\
				\hline
				\tabincell{l}{40m-1km \\ -8m/s} & \tabincell{l}{Tran. 0.516 \\ Vel. 0.1050}  & \tabincell{l}{Tran. \textbf{0.303} \\ Vel. 0.1042} & \tabincell{l}{Tran. 0.493 \\ Vel. \textbf{0.1006}} & \tabincell{l}{Tran. 5.694 \\ Vel. 0.4058}\\
                    \hline
                    Average & \tabincell{l}{Tran. \textbf{0.595} \\ Vel. \textbf{0.0905}}  & \tabincell{l}{Tran. 0.675 \\ Vel. 0.0909} & \tabincell{l}{Tran. 0.784 \\ Vel. 0.1017} & \tabincell{l}{Tran. 5.514 \\ Vel. 0.3301}\\
				\hline
			\end{tabular}
		\end{center}
	\end{table}

        \subsubsection{Dpsr and Ddcp Measurements}
        Once a base station is available, we can perform Dpsr [see Eq. (\ref{z_dp})] and Ddcp [see Eq. (\ref{z_ddcp})] measurements update. We conducted experiments on the following configurations: VIO tightly-coupled fusion with Dopp and Dpsr(VIO+Dopp+Dpsr), as well as VIO tightly-coupled fusion with Dopp and Ddcp(VIO+Dopp+Ddcp).

        The accuracy results are presented in Table \ref{dpsr_ddcp_table}. It is evident that the VIO+Dopp+Ddcp method achieved the highest position accuracy when the integer ambiguity was successfully resolved by the RTK module. When the ambiguity resolution was unsuccessful, the VIO+Dopp+Dpsr method still exhibited higher position accuracy compared to the VIO+Dopp+Psr method. In terms of velocity accuracy, the fusion of Psr, Dpsr, and Ddcp measurements on top of the VIO+Dopp fusion resulted in comparable levels of accuracy.

        We also performed an ablation experiment in which we excluded IMU integration for GNSS measurement update in the VIO+Dopp+Psr method. The results from Table \ref{dpsr_ddcp_table} indicate a significant decrease in both position and velocity accuracy when IMU integration was not utilized.
        
         \begin{table}[]
		\caption{RMSE with Dpsr and Ddcp measurements}
		\label{dpsr_ddcp_table}
		\begin{center}
			\begin{tabular}{c c c c c}
				\hline
				\tabincell{l}{Height-Len \\ -Vel} & \tabincell{l}{VIO+Dopp \\ +Ddcp} & \tabincell{l}{VIO+Dopp \\ +Dpsr} & \tabincell{l}{VIO+Dopp \\ +Psr} & \tabincell{l}{VIO+Dopp+ \\ Psr wo.inte}\\
				\hline
				\tabincell{l}{60m-1km \\ -6m/s} & \tabincell{l}{Tran. \textbf{0.118} \\ Vel. \textbf{0.0771}} & \tabincell{l}{Tran. 0.286 \\ Vel. 0.0782} & \tabincell{l}{Tran. 0.652 \\ Vel. 0.0813} & \tabincell{l}{Tran. 0.942 \\ Vel. 0.1316}\\
				\hline
				\tabincell{l}{60m-1km \\ -4m/s} & \tabincell{l}{Tran. \textbf{0.121} \\ Vel. 0.0683}  & \tabincell{l}{Tran. 0.306 \\ Vel. \textbf{0.0677}} & \tabincell{l}{Tran. 0.509 \\ Vel. 0.0706} & \tabincell{l}{Tran. 0.750 \\ Vel. 0.1136}\\
				\hline
				\tabincell{l}{60m-1km \\ -8m/s} & \tabincell{l}{Tran. \textbf{0.156} \\ Vel. 0.0910}  & \tabincell{l}{Tran. 0.344 \\ Vel. \textbf{0.0884}} & \tabincell{l}{Tran. 0.522 \\ Vel. 0.0933} & \tabincell{l}{Tran. 1.065 \\ Vel. 0.1486}\\
				\hline
				\tabincell{l}{60m-1km \\ -10m/s} & \tabincell{l}{Tran. \textbf{0.156} \\ Vel. 0.1013}  & \tabincell{l}{Tran. 0.419 \\ Vel. 0.0945} & \tabincell{l}{Tran. 0.642 \\ Vel. \textbf{0.0936}}& \tabincell{l}{Tran. 1.208 \\ Vel. 0.1646}\\
				\hline
				\tabincell{l}{60m-400m \\ -8m/s} & \tabincell{l}{Tran. \textbf{0.123} \\ Vel. 0.0672}  & \tabincell{l}{Tran. 0.254 \\ Vel. \textbf{0.0667}} & \tabincell{l}{Tran. 0.436 \\ Vel. 0.0690}& \tabincell{l}{Tran. 0.759 \\ Vel. 0.1257}\\
				\hline
				\tabincell{l}{50m-3km \\ -4m/s} & \tabincell{l}{Tran. \textbf{0.116} \\ Vel. 0.0779}  & \tabincell{l}{Tran. 0.387 \\ Vel. 0.0764} & \tabincell{l}{Tran. 0.707 \\ Vel. \textbf{0.0756}} & \tabincell{l}{Tran. 0.767 \\ Vel. 0.1027}\\
				\hline
				\tabincell{l}{50m-3km \\ -8m/s} & \tabincell{l}{Tran. \textbf{0.149} \\ Vel. 0.0942}  & \tabincell{l}{Tran. 0.742 \\ Vel. 0.0894} & \tabincell{l}{Tran. 0.804 \\ Vel. \textbf{0.0892}} & \tabincell{l}{Tran. 0.865 \\ Vel. 0.1401}\\
				\hline
				\tabincell{l}{50m-3km \\ -12m/s} & \tabincell{l}{Tran. \textbf{0.178} \\ Vel. \textbf{0.0930}}  & \tabincell{l}{Tran. 0.440 \\ Vel. 0.0982} & \tabincell{l}{Tran. \textbf{0.694} \\ Vel. 0.1006} & \tabincell{l}{Tran. 0.994 \\ Vel. 0.1808}\\
				\hline
				\tabincell{l}{20m-1km \\ -8m/s} & \tabincell{l}{Tran. \textbf{0.162} \\ Vel. 0.1355}  & \tabincell{l}{Tran. 0.485 \\ Vel. 0.1357} & \tabincell{l}{Tran. 0.467 \\ Vel. \textbf{0.1267}} & \tabincell{l}{Tran. 0.842 \\ Vel. 0.1535}\\
				\hline
				\tabincell{l}{40m-1km \\ -8m/s} & \tabincell{l}{Tran. \textbf{0.148} \\ Vel. \textbf{0.1041}}  & \tabincell{l}{Tran. 0.441 \\ Vel. 0.1067} & \tabincell{l}{Tran. 0.516 \\ Vel. 0.1050} & \tabincell{l}{Tran. 0.878 \\ Vel. 0.1458}\\
                    \hline
                    Average & \tabincell{l}{Tran. \textbf{0.143} \\ Vel. 0.0910}  & \tabincell{l}{Tran. 0.410 \\ Vel. \textbf{0.0902}} & \tabincell{l}{Tran. 0.595 \\ Vel. 0.0905} & \tabincell{l}{Tran. 0.907 \\ Vel. 0.1407}\\
				\hline
			\end{tabular}
		\end{center}
	\end{table}

	\subsection{Online GNSS-IMU Extrinsic Calibration}
        To investigate the impact of online GNSS-IMU extrinsic estimation, we conducted experiments using both accurate and poor initial GNSS-IMU extrinsic parameters. We also performed experiments with the GNSS-IMU extrinsic calibration enabled and disabled, separately.

        Table \ref{extrinsic_table} shows the accuracy results under different configurations. When given poor initial extrinsic parameters (with an error of 0.3m in the x-axis direction), enabling GNSS-IMU extrinsic estimation improved all velocity accuracy and almost all position accuracy. When given accurate initial extrinsic parameters, the effect of enabling GNSS-IMU extrinsic estimation was relatively minor, with only a slight improvement in average position and velocity accuracy. 

\begin{figure}[thpb]
		\centering
		\includegraphics[scale=0.25]{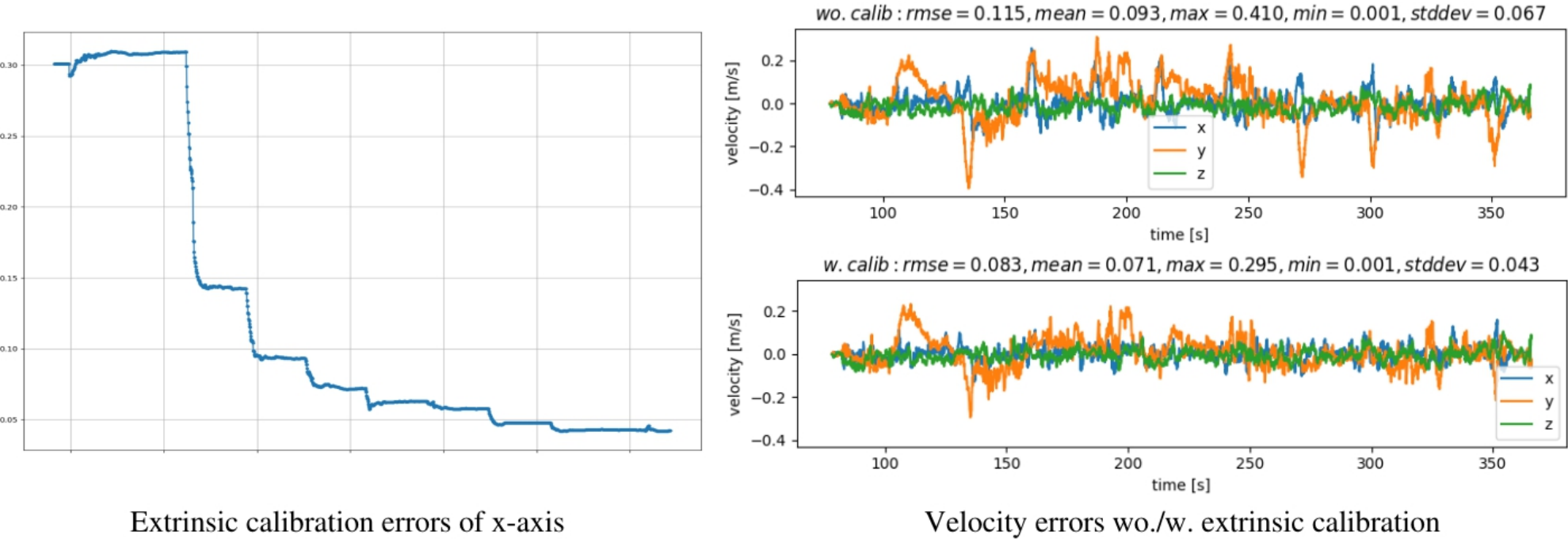}
		\caption{Estimation errors of GNSS-IMU extrinsic calibration and velocity errors in the 60m-1km-6m/s dataset with poor initial GNSS-IMU extrinsic.}
		\label{gnss_calib}
	\end{figure}

        Fig. \ref{gnss_calib} presents the estimation errors of GNSS-IMU extrinsic calibration when given poor initial GNSS-IMU extrinsic parameters. It can be observed that the estimation errors gradually converge to nearly zero. Additionally, the velocity errors with GNSS-IMU extrinsic calibration are significantly reduced compared to the case without extrinsic calibration. These results serve as validation that GNSS-IMU extrinsic estimation can effectively mitigate the impact of extrinsic parameter degradation and improve localization accuracy.

	\begin{table}[]
		\caption{Online GNSS-IMU extrinsic estimation results}
		\label{extrinsic_table}
		\begin{center}
			\begin{tabular}{c c c c c}
				\hline
				\tabincell{l}{Height-Len \\ -Vel} & \tabincell{l}{poor extr \\ + w.calib} & \tabincell{l}{poor extr \\ + wo.calib} & \tabincell{l}{good extr \\ + w.calib} & \tabincell{l}{good extr \\ + wo.calib} \\
				\hline
				\tabincell{l}{60m-1km \\ -6m/s} & \tabincell{l}{Tran. \textbf{0.622} \\ Vel. \textbf{0.0833}} & \tabincell{l}{Tran. 0.888 \\ Vel. 0.1150} & \tabincell{l}{Tran. 0.652 \\ Vel. \textbf{0.0813}} & \tabincell{l}{Tran. \textbf{0.650} \\ Vel. 0.0859}\\
				\hline
				\tabincell{l}{60m-1km \\ -4m/s} & \tabincell{l}{Tran. \textbf{0.476} \\ Vel. \textbf{0.0718}}  & \tabincell{l}{Tran. 0.690 \\ Vel. 0.1069} & \tabincell{l}{Tran. \textbf{0.509} \\ Vel. 0.0706} & \tabincell{l}{Tran. 0.518 \\ Vel. \textbf{0.0665}}\\
				\hline
				\tabincell{l}{60m-1km \\ -8m/s} & \tabincell{l}{Tran. \textbf{0.611} \\ Vel. \textbf{0.0958}}  & \tabincell{l}{Tran. 0.994 \\ Vel. 0.1328} & \tabincell{l}{Tran. 0.522 \\ Vel. 0.0933} & \tabincell{l}{Tran. \textbf{0.483} \\ Vel. \textbf{0.0913}} \\
				\hline
				\tabincell{l}{60m-1km \\ -10m/s} & \tabincell{l}{Tran. \textbf{0.642} \\ Vel. \textbf{0.0938}}  & \tabincell{l}{Tran. 1.083 \\ Vel. 0.1337} & \tabincell{l}{Tran. \textbf{0.642} \\ Vel. \textbf{0.0936}} & \tabincell{l}{Tran. 0.652 \\ Vel. 0.0965}\\
				\hline
				\tabincell{l}{60m-400m \\ -8m/s} & \tabincell{l}{Tran. \textbf{0.391} \\ Vel. \textbf{0.0751}}  & \tabincell{l}{Tran. 0.479 \\ Vel. 0.0963} & \tabincell{l}{Tran. 0.436 \\ Vel. \textbf{0.0690}} & \tabincell{l}{Tran. \textbf{0.426} \\ Vel. 0.0695}\\
				\hline
				\tabincell{l}{50m-3km \\ -4m/s} & \tabincell{l}{Tran. 0.781 \\ Vel. \textbf{0.0769}}  & \tabincell{l}{Tran. \textbf{0.654} \\ Vel. 0.1104} & \tabincell{l}{Tran. \textbf{0.707} \\ Vel. 0.0756} & \tabincell{l}{Tran. 0.717 \\ Vel. \textbf{0.0749}}\\
				\hline
				\tabincell{l}{50m-3km \\ -8m/s} & \tabincell{l}{Tran. 0.957 \\ Vel. \textbf{0.1120}}  & \tabincell{l}{Tran. \textbf{0.810} \\ Vel. 0.1403} & \tabincell{l}{Tran. \textbf{0.804} \\ Vel. \textbf{0.0892}} & \tabincell{l}{Tran. 0.945 \\ Vel. 0.0991}\\
				\hline
				\tabincell{l}{50m-3km \\ -12m/s} & \tabincell{l}{Tran. 0.707 \\ Vel. \textbf{0.0983}}  & \tabincell{l}{Tran. \textbf{0.585} \\ Vel. 0.1417} & \tabincell{l}{Tran. 0.694 \\ Vel. 0.1006} & \tabincell{l}{Tran. \textbf{0.646} \\ Vel. \textbf{0.0953}}\\
				\hline
				\tabincell{l}{20m-1km \\ -8m/s} & \tabincell{l}{Tran. \textbf{0.513} \\ Vel. \textbf{0.1296}}  & \tabincell{l}{Tran. 1.486 \\ Vel. 0.1671} & \tabincell{l}{Tran. \textbf{0.467} \\ Vel. 0.1267} & \tabincell{l}{Tran. 0.633 \\ Vel. \textbf{0.1239}}\\
				\hline
				\tabincell{l}{40m-1km \\ -8m/s} & \tabincell{l}{Tran. \textbf{0.475} \\ Vel. \textbf{0.1032}}  & \tabincell{l}{Tran. 0.777 \\ Vel. 0.1394} & \tabincell{l}{Tran. \textbf{0.516} \\ Vel. 0.1050} & \tabincell{l}{Tran. 0.536 \\ Vel. \textbf{0.1040}}\\
                    \hline
                    Average & \tabincell{l}{Tran. \textbf{0.618} \\ Vel. \textbf{0.0940}}  & \tabincell{l}{Tran. 0.844 \\ Vel. 0.1284} & \tabincell{l}{Tran. \textbf{0.595} \\ Vel. \textbf{0.0905}} & \tabincell{l}{Tran. 0.621 \\ Vel. 0.0907}\\
				\hline
			\end{tabular}
		\end{center}
	\end{table}

	\subsection{Real-Time Performance}
        Table \ref{timing_table} presents the CPU usage and run time of the VIO+Dopp+Dpsr method. We replayed the datasets on a Laptop with an Intel Core i7-10750U 2.60Ghz$\times$12 core CPU and 16GB memory, and conducted real-flight tests on a UAV platform featuring a Qualcomm RB5 processor. As shown in Table \ref{timing_table}, the fusion of Dopp and Dpsr measurements increases the estimator time by 13.77\% compared to the VIO method, accounting for 2.76\% of the total processing time. The CPU usage only increased from 22.77\% to 23.01\%, indicating that the computational cost brought by GNSS measurement update is almost negligible. Furthermore, the real-flight tests demonstrated that SRI-GVINS can achieve real-time performance even on resource-constrained devices.
	
	\begin{table}[]
		\caption{Timing results of SRI-GVINS on laptop and Qualcomm RB5}
		\label{timing_table}
		\begin{center}
			\begin{tabular}{l c c c c}
				\hline
				&  \tabincell{l}{Laptop \\ VIO} & \tabincell{l}{Laptop \\ SRI-GVINS} & \tabincell{l}{Qualcomm RB5 \\ SRI-GVINS} \\
				\hline
				CPU usage(\%)& 22.77 & 23.01 & 49.79\\
				\hline
				Estimator time(ms)& 2.230 & 2.537 & 6.129\\
				\hline
                    Total time(ms)& 8.525 & 8.760 & 34.739\\
				\hline
			\end{tabular}
		\end{center}
	\end{table}

        \section{EXPERIMENTS ON PUBLIC DATASETS}

\begin{figure}[thpb]
		\centering
		\includegraphics[scale=0.25]{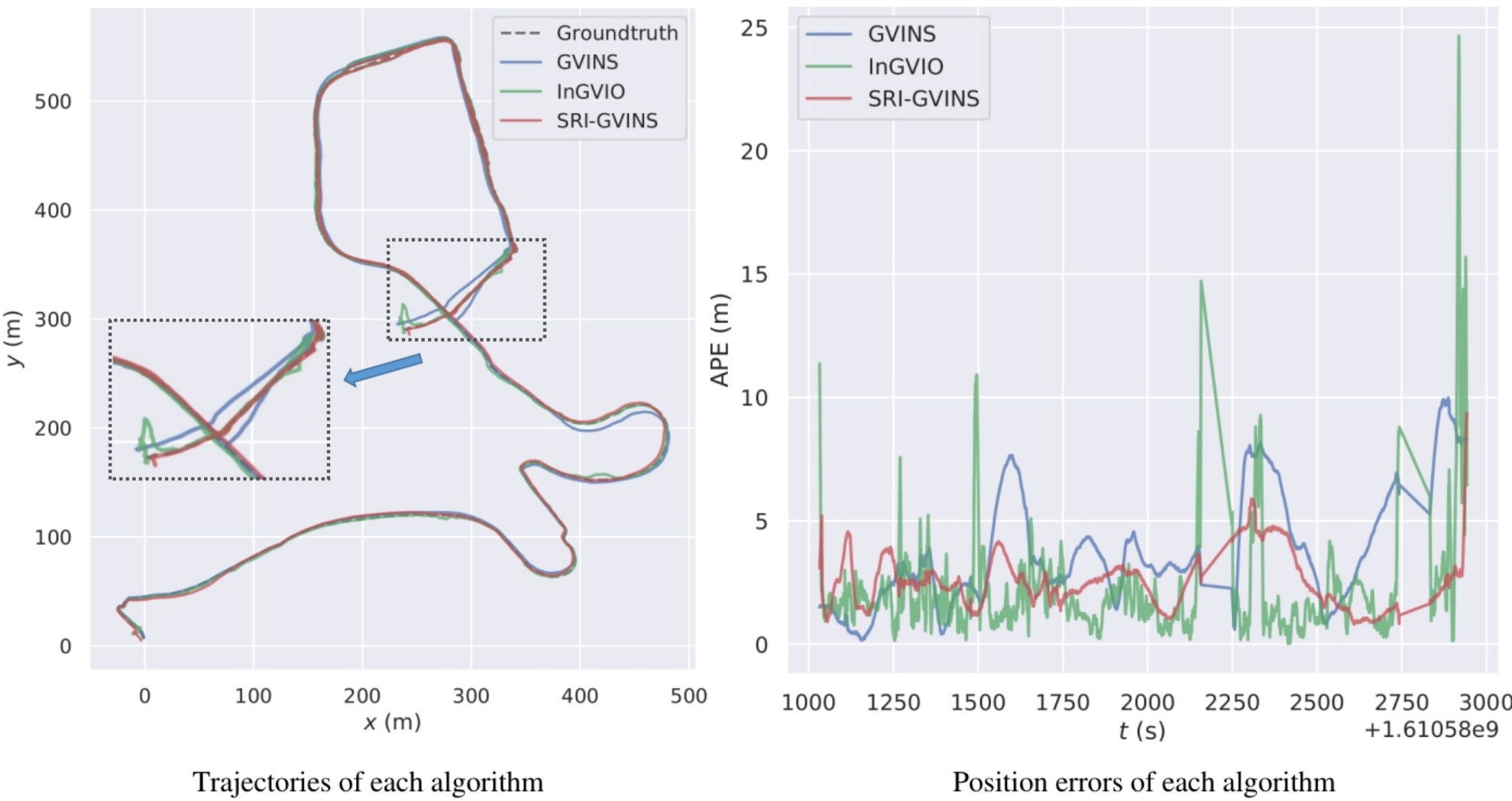}
		\caption{Trajectories and position errors of each algorithm in the complex\_environment dataset.}
		\label{gvins_evo}
	\end{figure}

        SRI-GVINS was evaluated and compared with two other state-of-the-art methods, GVINS \cite{cao2022gvins} and InGVIO \cite{liu2023ingvio}, using three public datasets provided by \cite{cao2022gvins}, which contain IMU, stereo images, Dopp and Psr measurenments. The comparison results are presented in Table \ref{gvins_table}, which shows the accuracy and efficiency of the algorithms. Fig. \ref{gvins_evo} shows the trajectories and position errors of the three algorithms, and it should be noted that any trajectory that is affected by unreliable RTK data has been excluded from the analysis. 
        
        From Table \ref{gvins_table}, SRI-GVINS exhibits competitive accuracy owing to its superior numerical stability and the GNSS measurement update with IMU integration. Moreover, SRI-GVINS achieves the lowest computational time, thanks to the efficiency gained through the utilization of the square-root form.

        \begin{table}[]
		\caption{Translational RMSE(m) and average time(ms) comparison}
		\label{gvins_table}
		\begin{center}
			\begin{tabular}{l c c c c}
				\hline
				&  GVINS & InGVIO & SRI-GVINS \\
				\hline
				sports\_field & \tabincell{l}{Trans. 0.578 \\ Time. 24.858} & \tabincell{l}{Trans. \textbf{0.566} \\ Time. 5.136} & \tabincell{l}{Trans. 0.717 \\ Time. \textbf{3.372}}\\
				\hline
				complex\_environment & \tabincell{l}{Trans. 4.376 \\ Time. 19.340} & \tabincell{l}{Trans. 3.063 \\ Time. 4.363} & \tabincell{l}{Trans. \textbf{2.783} \\ Time. \textbf{3.703}}\\
				\hline
                    urban\_driving & \tabincell{l}{Trans. - \\ Time. 24.858} & \tabincell{l}{Trans. 5.020 \\ Time. 3.202} & \tabincell{l}{Trans \textbf{4.548} \\ Time. \textbf{2.959}}\\
				\hline
			\end{tabular}
		\end{center}
	\end{table}

	\section{Conclusions and Future Work}
 
	In this paper, we have developed a tightly-coupled GNSS-Visual-Inertial navigation system that deeply fuses visual, inertial, and raw GNSS data (including pseudorange, Doppler shift, single-differenced pseudorange and double-differenced carrier phase) within the efficient SRI-SWF framework. We have used a filter-based initialization approach to adaptively converge the reference frame transformation, leveraged IMU integration to update the state using delayed and asynchronous GNSS measurement, and performed online calibration of GNSS-IMU extrinsic parameters. Extensive experimental results demonstrate that the proposed SRI-GVINS outperforms other state-of-the-art methods in terms of accuracy and efficiency.
	In the future, we will use raw GNSS measurements such as Doppler shift to assist in the initialization of the reference frame transform. Additionally, we will directly employ the LAMBDA algorithm \cite{temiissen1995least} to solve for the double-differenced ambiguity in SRI-GVINS. 
 We will also perform an observability analysis for SRI-GVINS to identify any potential degenerate motions.

        \section*{Acknowledgement}
        The authors would like to thank Yunjun Shen, Dan Su and Jingqi Jiang for their contributions to the development of the VIO framework based on the SRI-SWF. Additionally, appreciation is extended to Qian Zhang for her efforts in preprocessing the raw GNSS data.

	\bibliographystyle{ieeetr}
	\bibliography{mybib}

\begin{thebibliography}{10}

\bibitem{huang2019visual}
G.~Huang, ``Visual-inertial navigation: A concise review,'' in {\em 2019 international conference on robotics and automation (ICRA)}, pp.~9572--9582, IEEE, 2019.

\bibitem{qin2018vins}
T.~Qin, P.~Li, and S.~Shen, ``Vins-mono: A robust and versatile monocular visual-inertial state estimator,'' {\em IEEE Transactions on Robotics}, vol.~34, no.~4, pp.~1004--1020, 2018.

\bibitem{leutenegger2015keyframe}
S.~Leutenegger, S.~Lynen, M.~Bosse, R.~Siegwart, and P.~Furgale, ``Keyframe-based visual--inertial odometry using nonlinear optimization,'' {\em The International Journal of Robotics Research}, vol.~34, no.~3, pp.~314--334, 2015.

\bibitem{campos2021orb}
C.~Campos, R.~Elvira, J.~J.~G. Rodr{\'\i}guez, J.~M. Montiel, and J.~D. Tard{\'o}s, ``Orb-slam3: An accurate open-source library for visual, visual--inertial, and multimap slam,'' {\em IEEE Transactions on Robotics}, vol.~37, no.~6, pp.~1874--1890, 2021.

\bibitem{von2022dm}
L.~Von~Stumberg and D.~Cremers, ``Dm-vio: Delayed marginalization visual-inertial odometry,'' {\em IEEE Robotics and Automation Letters}, vol.~7, no.~2, pp.~1408--1415, 2022.

\bibitem{mourikis2007multi}
A.~I. Mourikis and S.~I. Roumeliotis, ``A multi-state constraint kalman filter for vision-aided inertial navigation,'' in {\em Proceedings 2007 IEEE international conference on robotics and automation}, pp.~3565--3572, IEEE, 2007.

\bibitem{li2013high}
M.~Li and A.~I. Mourikis, ``High-precision, consistent ekf-based visual-inertial odometry,'' {\em The International Journal of Robotics Research}, vol.~32, no.~6, pp.~690--711, 2013.

\bibitem{geneva2020openvins}
P.~Geneva, K.~Eckenhoff, W.~Lee, Y.~Yang, and G.~Huang, ``Openvins: A research platform for visual-inertial estimation,'' pp.~4666--4672, 2020.

\bibitem{kelly2011visual}
J.~Kelly and G.~S. Sukhatme, ``Visual-inertial sensor fusion: Localization, mapping and sensor-to-sensor self-calibration,'' {\em The International Journal of Robotics Research}, vol.~30, no.~1, pp.~56--79, 2011.

\bibitem{hesch2012observability}
J.~A. Hesch, D.~G. Kottas, S.~L. Bowman, and S.~I. Roumeliotis, ``Observability-constrained vision-aided inertial navigation,'' {\em University of Minnesota, Dept. of Comp. Sci. \& Eng., MARS Lab, Tech. Rep}, vol.~1, p.~6, 2012.

\bibitem{angelino2012uav}
C.~V. Angelino, V.~R. Baraniello, and L.~Cicala, ``Uav position and attitude estimation using imu, gnss and camera,'' in {\em 2012 15th International Conference on Information Fusion}, pp.~735--742, IEEE, 2012.

\bibitem{lynen2013robust}
S.~Lynen, M.~W. Achtelik, S.~Weiss, M.~Chli, and R.~Siegwart, ``A robust and modular multi-sensor fusion approach applied to mav navigation,'' in {\em 2013 IEEE/RSJ international conference on intelligent robots and systems}, pp.~3923--3929, IEEE, 2013.

\bibitem{shen2014multi}
S.~Shen, Y.~Mulgaonkar, N.~Michael, and V.~Kumar, ``Multi-sensor fusion for robust autonomous flight in indoor and outdoor environments with a rotorcraft mav,'' in {\em 2014 IEEE International Conference on Robotics and Automation (ICRA)}, pp.~4974--4981, IEEE, 2014.

\bibitem{schreiber2016vehicle}
M.~Schreiber, H.~K{\"o}nigshof, A.-M. Hellmund, and C.~Stiller, ``Vehicle localization with tightly coupled gnss and visual odometry,'' in {\em 2016 IEEE Intelligent Vehicles Symposium (IV)}, pp.~858--863, IEEE, 2016.

\bibitem{mascaro2018gomsf}
R.~Mascaro, L.~Teixeira, T.~Hinzmann, R.~Siegwart, and M.~Chli, ``Gomsf: Graph-optimization based multi-sensor fusion for robust uav pose estimation,'' in {\em 2018 IEEE International Conference on Robotics and Automation (ICRA)}, pp.~1421--1428, IEEE, 2018.

\bibitem{yu2019gps}
Y.~Yu, W.~Gao, C.~Liu, S.~Shen, and M.~Liu, ``A gps-aided omnidirectional visual-inertial state estimator in ubiquitous environments,'' in {\em 2019 IEEE/RSJ International Conference on Intelligent Robots and Systems (IROS)}, pp.~7750--7755, IEEE, 2019.

\bibitem{qin2019general}
T.~Qin, S.~Cao, J.~Pan, and S.~Shen, ``A general optimization-based framework for global pose estimation with multiple sensors,'' {\em arXiv preprint arXiv:1901.03642}, 2019.

\bibitem{gong2020graph}
Z.~Gong, P.~Liu, F.~Wen, R.~Ying, X.~Ji, R.~Miao, and W.~Xue, ``Graph-based adaptive fusion of gnss and vio under intermittent gnss-degraded environment,'' {\em IEEE Transactions on Instrumentation and Measurement}, vol.~70, pp.~1--16, 2020.

\bibitem{lee2020intermittent}
W.~Lee, K.~Eckenhoff, P.~Geneva, and G.~Huang, ``Intermittent gps-aided vio: Online initialization and calibration,'' in {\em 2020 IEEE International Conference on Robotics and Automation (ICRA)}, pp.~5724--5731, IEEE, 2020.

\bibitem{xiong2021g}
L.~Xiong, R.~Kang, J.~Zhao, P.~Zhang, M.~Xu, R.~Ju, C.~Ye, and T.~Feng, ``G-vido: A vehicle dynamics and intermittent gnss-aided visual-inertial state estimator for autonomous driving,'' {\em IEEE Transactions on Intelligent Transportation Systems}, vol.~23, no.~8, pp.~11845--11861, 2021.

\bibitem{won2014gnss}
D.~H. Won, E.~Lee, M.~Heo, S.~Sung, J.~Lee, and Y.~J. Lee, ``Gnss integration with vision-based navigation for low gnss visibility conditions,'' {\em GPS solutions}, vol.~18, pp.~177--187, 2014.

\bibitem{li2019tight}
T.~Li, H.~Zhang, Z.~Gao, X.~Niu, and N.~El-Sheimy, ``Tight fusion of a monocular camera, mems-imu, and single-frequency multi-gnss rtk for precise navigation in gnss-challenged environments,'' {\em Remote Sensing}, vol.~11, no.~6, p.~610, 2019.

\bibitem{cao2022gvins}
S.~Cao, X.~Lu, and S.~Shen, ``Gvins: Tightly coupled gnss--visual--inertial fusion for smooth and consistent state estimation,'' {\em IEEE Transactions on Robotics}, vol.~38, no.~4, pp.~2004--2021, 2022.

\bibitem{liu2021optimization}
J.~Liu, W.~Gao, and Z.~Hu, ``Optimization-based visual-inertial slam tightly coupled with raw gnss measurements,'' in {\em 2021 IEEE International Conference on Robotics and Automation (ICRA)}, pp.~11612--11618, IEEE, 2021.

\bibitem{li2022p}
T.~Li, L.~Pei, Y.~Xiang, W.~Yu, and T.-K. Truong, ``P3-vins: Tightly-coupled ppp/ins/visual slam based on optimization approach,'' {\em IEEE Robotics and Automation Letters}, vol.~7, no.~3, pp.~7021--7027, 2022.

\bibitem{lee2022tightly}
W.~Lee, P.~Geneva, Y.~Yang, and G.~Huang, ``Tightly-coupled gnss-aided visual-inertial localization,'' in {\em 2022 International Conference on Robotics and Automation (ICRA)}, pp.~9484--9491, IEEE, 2022.

\bibitem{liu2022variable}
C.~Liu, C.~Jiang, and H.~Wang, ``Variable observability constrained visual-inertial-gnss ekf-based navigation,'' {\em IEEE Robotics and Automation Letters}, vol.~7, no.~3, pp.~6677--6684, 2022.

\bibitem{liu2023ingvio}
C.~Liu, C.~Jiang, and H.~Wang, ``Ingvio: A consistent invariant filter for fast and high-accuracy gnss-visual-inertial odometry,'' {\em IEEE Robotics and Automation Letters}, vol.~8, no.~3, pp.~1850--1857, 2023.

\bibitem{maybeck1982stochastic}
P.~S. Maybeck, {\em Stochastic models, estimation, and control}.
\newblock Academic press, 1982.

\bibitem{bierman2006factorization}
G.~J. Bierman, {\em Factorization methods for discrete sequential estimation}.
\newblock Courier Corporation, 2006.

\bibitem{wu2015square}
K.~Wu, A.~M. Ahmed, G.~A. Georgiou, and S.~I. Roumeliotis, ``A square root inverse filter for efficient vision-aided inertial navigation on mobile devices.,'' in {\em Robotics: Science and Systems}, vol.~2, p.~2, Rome, Italy, 2015.

\bibitem{civera2008inverse}
J.~Civera, A.~J. Davison, and J.~M. Montiel, ``Inverse depth parametrization for monocular slam,'' {\em IEEE transactions on robotics}, vol.~24, no.~5, pp.~932--945, 2008.

\bibitem{trawny2005indirect}
N.~Trawny and S.~I. Roumeliotis, ``Indirect kalman filter for 3d attitude estimation,'' {\em University of Minnesota, Dept. of Comp. Sci. \& Eng., Tech. Rep}, vol.~2, p.~2005, 2005.

\bibitem{fischler1981random}
M.~A. Fischler and R.~C. Bolles, ``Random sample consensus: a paradigm for model fitting with applications to image analysis and automated cartography,'' {\em Communications of the ACM}, vol.~24, no.~6, pp.~381--395, 1981.

\bibitem{li2014visual}
M.~Li, {\em Visual-inertial odometry on resource-constrained systems}.
\newblock University of California, Riverside, 2014.

\bibitem{forster2016manifold}
C.~Forster, L.~Carlone, F.~Dellaert, and D.~Scaramuzza, ``On-manifold preintegration for real-time visual--inertial odometry,'' {\em IEEE Transactions on Robotics}, vol.~33, no.~1, pp.~1--21, 2016.

\bibitem{sturm2012benchmark}
J.~Sturm, N.~Engelhard, F.~Endres, W.~Burgard, and D.~Cremers, ``A benchmark for the evaluation of rgb-d slam systems,'' in {\em 2012 IEEE/RSJ international conference on intelligent robots and systems}, pp.~573--580, IEEE, 2012.

\bibitem{temiissen1995least}
J.~Temiissen, ``The least-squares ambiguity decorrelation adjustment: A method for fast gps integer ambiguity estimation,'' {\em Journal of Geodesy}, vol.~70, pp.~65--82, 1995.

\end{thebibliography}

\end{document}